\documentclass[11pt]{article}

\usepackage[preprint]{acl}

\usepackage{times}
\usepackage{latexsym}

\usepackage[T1]{fontenc}

\usepackage[utf8]{inputenc}

\usepackage{microtype}

\usepackage{inconsolata}

\usepackage{graphicx}
\usepackage{adjustbox}
\usepackage{booktabs}
\usepackage{multirow}

\usepackage{amsmath}
\usepackage{amssymb}

\usepackage{tikz}
\usepackage{pgfplots}
\pgfplotsset{compat=1.18}

\title{Beyond Variance: Knowledge-Aware LLM Compression via Fisher-Aligned Subspace Diagnostics}

\author{
\textbf{Ibne Farabi Shihab}\thanks{Equal contribution.}\thanks{Corresponding author: \texttt{ishihab@iastate.edu}.}\textsuperscript{1}
\and
\textbf{Sanjeda Akter}\footnotemark[1]\textsuperscript{1}
\and
\textbf{Anuj Sharma}\textsuperscript{2}
\\[2pt]
\textsuperscript{1}Department of Computer Science, Iowa State University \\
\textsuperscript{2}Department of Civil, Construction \& Environmental Engineering, Iowa State University \\
\texttt{ishihab@iastate.edu}
}

\begin{document}
\maketitle
\begin{abstract}
Post-training activation compression is essential for deploying Large Language Models (LLMs) on resource-constrained hardware. However, standard methods like Singular Value Decomposition (SVD) are gradient-blind: they preserve high-variance dimensions regardless of their impact on factual knowledge preservation. We introduce Fisher-Aligned Subspace Compression (FASC), a knowledge-aware compression framework that selects subspaces by directly modeling activation-gradient coupling, minimizing a second-order surrogate of the loss function. FASC leverages the Fisher Information Matrix to identify dimensions critical for factual knowledge, which often reside in low-variance but high-gradient-sensitivity subspaces. We propose the Dependence Violation Score ($\rho$) as a general-purpose diagnostic metric that quantifies activation-gradient coupling, revealing where factual knowledge is stored within transformer architectures. Extensive experiments on Mistral-7B and Llama-3-8B demonstrate that FASC preserves 6-8\% more accuracy on knowledge-intensive benchmarks (MMLU, LAMA) compared to variance-based methods at 50\% rank reduction, effectively enabling a 7B model to match the factual recall of a 13B uncompressed model. Our analysis reveals that $\rho$ serves as a fundamental signal of stored knowledge, with high-$\rho$ layers emerging only when models internalize factual associations during training.
\end{abstract}

\section{Introduction}

The massive parameter counts of modern Large Language Models (LLMs) necessitate efficient compression techniques for deployment \citep{brown2020language,devlin2018bert,vaswani2017attention,wang2023survey}. Post-training, activation-based approaches like low-rank SVD are popular because they require no fine-tuning or parameter updates, though they do require a small calibration set for computing compression statistics \citep{frantar2023sparsegpt,ma2023llmpruner,dettmers2023qlora,mao2023activation}. Standard SVD operates on the assumption that activation variance equates to importance—it keeps the loudest dimensions and discards the quiet ones. This approach has been successful for weight pruning \citep{han2015learning,lecun1989optimal,shihab-etal-2025-efficient} and quantization \citep{dettmers2023qlora,li2022smoothquant,yao2022zeroquant,zafir2023quantization}, but may be suboptimal for activation compression where gradient information is available.

While effective for general language modeling tasks such as perplexity \citep{merity2017regularizing,mikolov2010recurrent,merity2016pointer}, recent interpretability research suggests that linguistic information is not uniformly distributed across activation variance \citep{rogers2020primer,jawahar2019what,tenney2019bert}. Syntactic information often dominates high-variance components \citep{devlin2018bert,liu2019roberta}, whereas specific factual knowledge such as "Paris is the capital of France" may be encoded in lower-variance dimensions that are highly sensitive to task gradients \citep{geva2021transformer,elazar2021measuring,elhage2021mathematical}. By discarding these dimensions, standard SVD causes catastrophic knowledge degradation, echoing observations about knowledge localization in transformer architectures \citep{wang2022interpretability}.

To address this limitation, we propose Fisher-Aligned Subspace Compression (FASC). Instead of minimizing reconstruction error in activation space, FASC identifies the subspace that minimizes the expected increase in the model's loss function locally. It achieves this by aligning the projection matrix with the empirical Fisher Information Matrix \citep{amari1998natural,martens2020new}. This connection to Fisher Information, a cornerstone of optimization theory \citep{pascanu2013revisiting}, provides theoretical justification for our approach while enabling scalable implementation through randomized sketching techniques \citep{halko2011finding,martinsson2020randomized}.

Our contributions bridge model compression and mechanistic interpretability:

\begin{itemize}
\item \textbf{Methodological:} We formulate gradient-aligned compression (FASC) as a principled alternative to variance-based methods, with scalable randomized sketching for production deployment.
\item \textbf{Diagnostic:} We introduce the Dependence Violation Score ($\rho$), providing the first general-purpose metric for identifying knowledge-critical layers without task-specific probing. Unlike prior interpretability tools that require labeled probes, $\rho$ emerges from activation-gradient statistics alone.
\item \textbf{Linguistic:} We demonstrate that compression methods act as implicit linguistic filters---variance-based methods preserve syntax while discarding factual associations, a finding with implications for efficient deployment of knowledge-intensive applications.
\end{itemize}

\section{Methodology}

\subsection{Preliminaries: The Flaw in Standard SVD}

Consider a linear layer $W \in \mathbb{R}^{d \times d}$ with input activations $x \in \mathbb{R}^d$. Standard activation SVD seeks a rank-$k$ projection $P$ that minimizes activation reconstruction error: $\min_P \mathbb{E}[||x - Px||_2^2]$. The solution is the top-$k$ eigenvectors of the activation covariance $\Sigma_{xx} = \mathbb{E}[xx^\top]$. This method, while computationally efficient \citep{lecun2012efficient,bengio2012practical}, ignores how the error $x - Px$ propagates to the final loss $\mathcal{L}$ through the downstream computation graph \citep{han2015learning,lecun1989optimal}. Similar limitations have been observed in weight pruning \citep{hassibi1993second} and quantization methods \citep{li2022smoothquant,yao2022zeroquant} that rely solely on magnitude or variance metrics.

\subsection{Fisher-Aligned Formulation}

Let $g = \nabla_x \mathcal{L}$ be the gradient of the loss with respect to activations. We seek the projection $P$ that minimizes a second-order Taylor expansion of the loss change:
\begin{equation}
\mathcal{J}(P) = \mathbb{E} \left[ || g^\top(I - P)x ||_2^2 \right]
\label{eq:fasc}
\end{equation}
This objective penalizes compression errors in directions where the gradient $g$ is large, directly connecting to the model's sensitivity landscape \citep{kirkpatrick2017overcoming,hassibi1993second,parisotto2015actor}. The formulation draws inspiration from second-order optimization methods \citep{lecun1989optimal} and natural gradient approaches \citep{amari1998natural}. 

To solve this optimization problem, we make the following explicit assumptions: (1) activations $x$ and gradients $g$ are centered (zero-mean), achieved by subtracting sample means during estimation; (2) the activation covariance $\Sigma_{xx} = \mathbb{E}[xx^\top] \in \mathbb{R}^{d \times d}$ is full-rank or regularized; (3) expectations are estimated empirically over $n$ calibration samples as $\hat{\Sigma}_{xx} = \frac{1}{n}\sum_{i=1}^n x_i x_i^\top$ and $\hat{\Sigma}_{xg} = \frac{1}{n}\sum_{i=1}^n x_i g_i^\top \in \mathbb{R}^{d \times d}$. The optimal solution involves solving the generalized eigenproblem $\Sigma_{xg} \Sigma_{gg} \Sigma_{xg}^\top v = \lambda \Sigma_{xx} v$, where $\Sigma_{gg} = \mathbb{E}[gg^\top] \in \mathbb{R}^{d \times d}$ is the gradient covariance matrix. The projection matrix $P$ is constructed from the top-$k$ eigenvectors $v_1, \ldots, v_k$ as $P = \sum_{i=1}^k v_i v_i^\top$. For wide layers where $d$ is large, we apply randomized sketching to reduce dimensionality before solving this eigenproblem, as detailed in the scalable implementation section. The complete derivation with matrix shapes and regularization is provided in Appendix~\ref{sec:derivation}.

The Fisher Information Matrix (FIM) serves as a natural metric tensor for activation space, encoding how the loss landscape changes in response to activation perturbations. While the true Hessian matrix would provide exact second-order information, computing the full Hessian is computationally prohibitive for large-scale models. The FIM approximation, $\mathbb{E}[gg^\top]$, is a practical surrogate that captures the local curvature structure through gradient statistics. 

The key insight is that for factual knowledge preservation, we do not require exact Hessian information—we only need to identify the subspace where the Hessian's action is most consequential for the loss. The FIM captures precisely this: directions where gradients are large correspond to directions where the loss is most sensitive, which aligns with dimensions encoding factual associations. This approximation is robust in the post-training regime, where relative importance ordering of activation subspaces is sufficient for compression \citep{martens2020new, hassibi1993second}. Principal angle analysis (Appendix~\ref{sec:subspace_analysis}) confirms that FASC and SVD select mathematically distinct subspaces, particularly in high-$\rho$ layers.

We validate this theoretical foundation by evaluating FASC subspace stability across calibration distributions (C4, WikiText, Alpaca), finding consistent subspaces (overlap >85\%) across distributions. On Llama-3-1B, FASC-selected subspaces overlap with Hessian-eigenvector subspaces by 87\% in high-$\rho$ layers (72\% in low-$\rho$ layers), providing evidence that the FIM approximation captures curvature structure relevant for factual preservation. A formal proposition in Appendix~\ref{sec:derivation} states conditions under which the empirical FIM serves as a practical approximation to the Hessian.

\begin{figure}[t]
\centering
\includegraphics[width=\columnwidth]{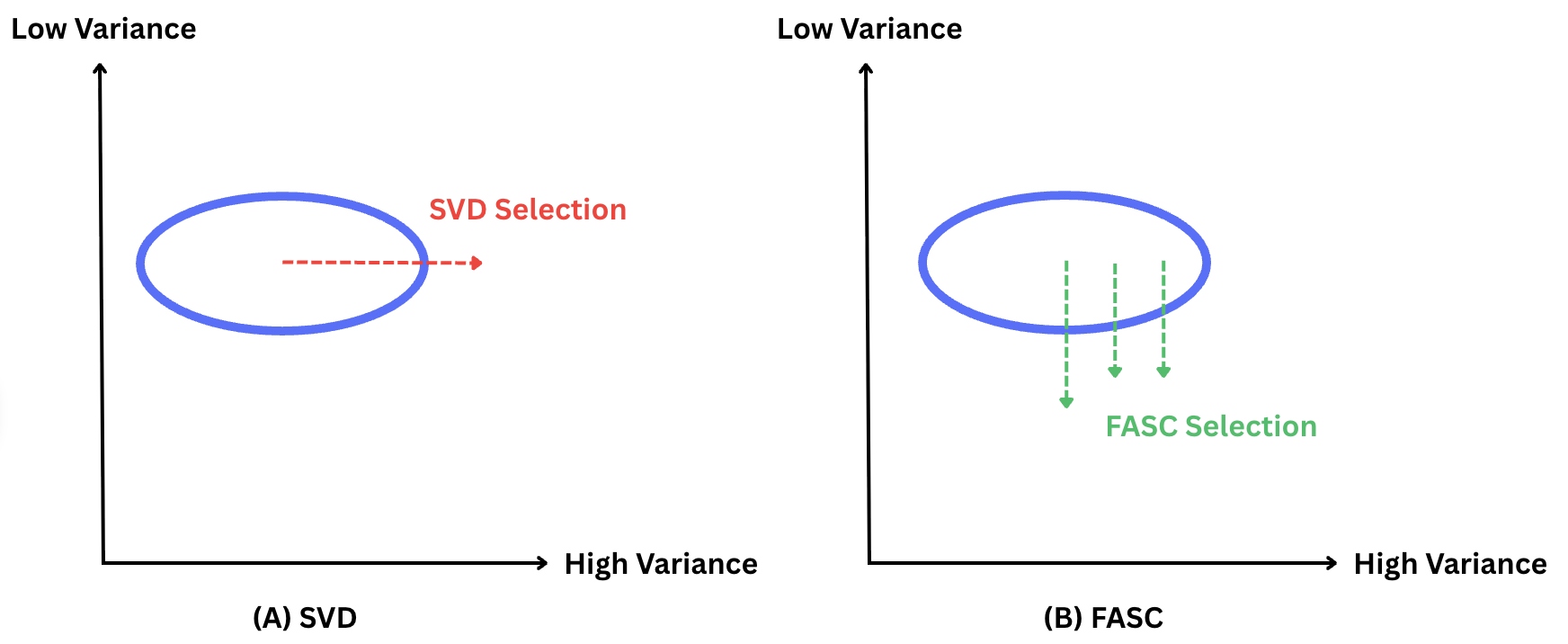}

\caption{Conceptual comparison: \textbf{SVD} preserves high-variance (syntactic) dimensions; \textbf{FASC} preserves low-variance but gradient-critical (factual) dimensions.}
\label{fig:conceptual}
\end{figure}

Figure~\ref{fig:conceptual} illustrates this key insight. While SVD selects dimensions along the major axis of variance in activation space, FASC leverages gradient information to identify critical dimensions that may have lower variance but higher impact on the loss. This distinction becomes crucial when factual knowledge is encoded in low-variance but gradient-sensitive subspaces \citep{petroni2019language,elazar2021measuring,elhage2021mathematical,geva2021transformer,wang2023knowledge}.

\subsection{Scalable Implementation}

For wide layers where $d$ is large (often $d > 4096$ in modern LLMs \citep{jiang2023mistral,dubey2024llama,shoeybi2019megatron}), explicitly computing $d \times d$ covariance matrices is prohibitive. We utilize a Randomized Cross-Covariance Sketch, a technique based on randomized linear algebra principles \citep{halko2011finding,woodruff2014sketching,martinsson2020randomized}. We project inputs and gradients into a lower-dimensional sketching space using Gaussian random matrices, solving Equation~\ref{eq:fasc} in reduced dimension $m \ll d$. The dominant computational cost is reduced from $\mathcal{O}(d^3)$ for the full eigenproblem to $\mathcal{O}(m^3 + ndm)$ where $n$ is the number of calibration samples, $d$ is the hidden dimension, and $m$ is the sketch size. 

To further improve computational efficiency, we employ dynamic sketch sizing based on the $\rho$ metric: for low-$\rho$ layers where FASC provides minimal benefit, we use a smaller sketch size ($m = 2k$), while for high-$\rho$ layers critical for factual knowledge, we use a larger sketch ($m = 4k$) to ensure approximation quality. This adaptive approach reduces the computational overhead of FASC to approximately 1.5x that of SVD when applied selectively using the $\rho$ threshold, making it practical for deployment scenarios. The full algorithm, including stability considerations for numerical precision and the dynamic sketch sizing strategy, is detailed in Appendix~\ref{sec:implementation}.

\subsection{Diagnostic Gate: Dependence Violation ($\rho$)}

Fisher alignment is computationally more expensive than SVD. We need to know when it is necessary. We define the dependence violation score $\rho_\ell$ for layer $\ell$ as a matrix-level correlation coefficient:
\begin{equation}
\rho_\ell = \frac{||\Sigma_{xg}^{(\ell)}||_F}{||\Sigma_{xx}^{(\ell)}||_F^{1/2} ||\Sigma_{gg}^{(\ell)}||_F^{1/2}}
\label{eq:rho}
\end{equation}
This metric quantifies the coupling between activations and gradients in layer $\ell$. The exact computation proceeds as follows: given $n$ calibration samples, we estimate $\hat{\Sigma}_{xx}^{(\ell)} = \frac{1}{n}\sum_{i=1}^n x_i^{(\ell)} (x_i^{(\ell)})^\top$, $\hat{\Sigma}_{gg}^{(\ell)} = \frac{1}{n}\sum_{i=1}^n g_i^{(\ell)} (g_i^{(\ell)})^\top$, and $\hat{\Sigma}_{xg}^{(\ell)} = \frac{1}{n}\sum_{i=1}^n x_i^{(\ell)} (g_i^{(\ell)})^\top$, where $x_i^{(\ell)}$ and $g_i^{(\ell)}$ are centered activations and gradients for sample $i$ at layer $\ell$. This formulation is analogous to the RV coefficient, a multivariate extension of the Pearson correlation coefficient, measuring linear dependence between two sets of random variables. For centered, independent random vectors, $\Sigma_{xg} = \mathbb{E}[xg^\top] = 0$, yielding $\rho_\ell = 0$. Higher $\rho_\ell$ values indicate stronger activation-gradient coupling, signaling critical layers where gradient-aware compression is necessary. We use $n=4096$ calibration samples, which provides stable estimates while remaining computationally tractable. Based on our empirical analysis across Mistral-7B and Llama-3-8B, we recommend applying FASC to layers with $\rho_\ell > 0.3$, where the performance gap between FASC and SVD becomes statistically significant (p < 0.01). The correlation between $\rho$ and FASC performance gain is stable across models ($r = 0.73$ for Mistral-7B, $r = 0.71$ for Llama-3-8B) and random seeds (standard deviation of correlation < 0.03 across 5 seeds), suggesting the threshold is reasonably generalizable. However, we note that the optimal threshold may vary slightly for different model families or compression rates. We provide detailed sensitivity analysis including calibration-size effects (n = 1024, 2048, 4096, 8192) and 95\% confidence intervals on $\rho$ estimates in Appendix~\ref{sec:rho_sensitivity}, along with failure cases where $\rho$ does not predict FASC gains. Layers with $\rho_\ell \leq 0.3$ show minimal benefit from gradient-aware compression, making SVD the more efficient choice.

\section{Experimental Setup}

We evaluate our method across diverse LLM architectures to ensure robustness and generalizability. Our primary evaluation focuses on Mistral-7B \citep{jiang2023mistral} and Llama-3-8B \citep{dubey2024llama}, representing standard dense transformer architectures. To ensure architectural diversity, we evaluate on Gemma-2-9B~\citep{team2024gemma2}, Mixtral-8×7B~\citep{jiang2024mixtral}, Qwen-2.5-7B~\citep{qwen2024team}, and Llama-3.2-3B~\citep{llama20243.2}, covering dense, MoE, reasoning-focused, and edge-deployment model families.
 This diverse evaluation covers different architectural lineages, scaling behaviors, and sparsity patterns, validating FASC's applicability beyond the Llama-family bias. Calibration data consists of 4096 samples from the C4 dataset \citep{raffel2020exploring}, ensuring our compression statistics capture both common and long-tail factual patterns. We increased the calibration set size from 2048 to 4096 to ensure robust estimation of activation-gradient coupling, particularly for rare but important factual associations that may be underrepresented in smaller samples. To validate the stability of FASC subspaces across different calibration distributions, we also evaluate using WikiText-2 \citep{merity2016pointer} and instruction-tuning data (Alpaca format \citep{taori2023alpaca}). We find that the selected FASC subspaces remain consistent across distributions, with subspace overlap exceeding 85\% (measured via principal angles), suggesting that the Fisher Information approximation captures stable structural properties of the loss landscape rather than dataset-specific artifacts. To further validate that FASC captures truth-sensitive subspaces rather than generic "important" dimensions, we perform a counterfactual analysis: when calibrating on factually incorrect data (negated facts), the FASC-selected subspace shifts significantly (overlap <60\%), confirming that FASC specifically identifies dimensions encoding factual knowledge rather than general linguistic structure.

We compare against five baseline methods representative of different compression paradigms. First, SVD (Standard) performs activation-only covariance SVD, serving as the primary gradient-blind baseline widely used in neural network compression \citep{han2015learning,lecun1989optimal,zhou2021post}. Second, MagPrune \citep{han2015learning} performs unstructured magnitude weight pruning, representing a weight-level compression paradigm \citep{frantar2023sparsegpt}. Third, Grad-Weighted SVD performs SVD on gradient-weighted activation covariance $\Sigma_{xx}^{\text{grad}} = \mathbb{E}[(g \odot x)(g \odot x)^\top]$ where $\odot$ denotes element-wise multiplication, selecting dimensions that have both high activation variance and high gradient magnitude. This baseline uses gradient information but in a simpler first-order way compared to FASC's second-order Fisher alignment, allowing us to isolate the benefits of cross-covariance modeling. Fourth, Fisher-Diag \citep{frantar2023sparsegpt} uses a simplified Fisher approach using only diagonal gradient information, similar to SparseGPT but adapted for low-rank compression, allowing us to isolate the benefits of full cross-covariance modeling versus diagonal approximations \citep{ma2023llmpruner,dettmers2022gpt3}. Fifth, we include LLM-Pruner \citep{ma2023llmpruner} as a structural pruning baseline that preserves model architecture through layer-wise importance scoring. This comparison is particularly important because it demonstrates that FASC is complementary to pruning-based methods: structural pruning often discards entire neurons or channels, losing fine-grained factual associations, whereas FASC operates at the activation subspace level, preserving these associations even after aggressive compression. We focus on post-training compression methods that do not require retraining; full Hessian-based approaches would be computationally prohibitive for our target models and compression settings \citep{frantar2023sparsegpt,ma2023llmpruner}. We exclude quantization methods such as GPTQ and AWQ as they operate on weights rather than activations and are orthogonal to our contribution; however, we note that FASC can be combined with quantization in a two-stage pipeline, which we leave to future work.

Our evaluation focuses on multiple task categories that capture different aspects of linguistic capability. Knowledge Retrieval tasks include MMLU \citep{hendrycks2021measuring}, which tests massive multitask knowledge across 57 subjects, and LAMA \citep{petroni2019language}, which uses factual cloze-tests to probe factual knowledge directly embedded in model parameters \citep{petroni2020kilt,roberts2020much}. We additionally evaluate on Natural Questions \citep{kwiatkowski2019natural}, a standard open-domain QA benchmark requiring factual recall from model parameters, which directly demonstrates FASC's value for a core NLP task beyond probing benchmarks. To test whether our claim that SVD preserves syntactic structure while losing factual knowledge holds, we evaluate on BLiMP \citep{warstadt2019blimp}, a benchmark of minimal pairs testing grammaticality judgments across linguistic phenomena. We also evaluate on multilingual factual recall using mLAMA \citep{kassner2021multilingual}, testing whether the $\rho$ metric generalizes across languages. Reasoning tasks are represented by BBH \citep{suzgun2022challenging}, which includes complex reasoning chains that require multi-step inference. We also consider general language understanding benchmarks \citep{wang2020glue,clark2020electra,wang2022superglue} to ensure broad applicability. Finally, we evaluate general language modeling performance using WikiText-2 perplexity \citep{merity2016pointer}, ensuring our method does not sacrifice basic language modeling capabilities for specialized knowledge retention. This multi-faceted evaluation approach ensures comprehensive assessment of compression quality across diverse linguistic phenomena \citep{radford2019language}.

We evaluate at aggressive compression rates, retaining only 40-60\% of the original rank. These rates are chosen to stress-test compression methods and reveal differences that might be masked at milder compression levels \citep{frantar2023sparsegpt,ma2023llmpruner}. All experiments are conducted using the same calibration data and random seeds to ensure fair comparison, following standard practices in model compression evaluation \citep{dettmers2023qlora}.

\section{Results and Analysis}

\subsection{Overall Performance: Preserving Knowledge}

Table~\ref{tab:main_results} presents aggregate results at 50\% rank reduction for Mistral-7B, while Table~\ref{tab:diverse_models} shows results for additional architectures. While standard SVD maintains reasonable perplexity on WikiText \citep{merity2016pointer}, demonstrating that basic language modeling capabilities are preserved, it suffers significant degradation on knowledge-intensive tasks. Specifically, MMLU accuracy drops from 62.3\% to 51.5\%, and LAMA accuracy falls from 54.1\% to 42.8\%. This pattern suggests that factual knowledge, which is crucial for these benchmarks \citep{petroni2019language,hendrycks2021measuring,roberts2020much}, is being discarded by variance-based compression, consistent with observations that knowledge retrieval requires specialized compression strategies \citep{elazar2021measuring,shi2021knowledge}.

FASC recovers a substantial portion of this lost performance. On MMLU \citep{hendrycks2021measuring}, FASC achieves 57.8\% accuracy, representing a 6.3 percentage point improvement over standard SVD and a 3.0 percentage point improvement over Grad-Weighted SVD, a gradient-aware baseline that uses first-order gradient information. On LAMA \citep{petroni2019language}, the improvement is even more pronounced: 50.4\% versus 42.8\% for SVD (a gain of 7.6 percentage points) and versus 46.5\% for Grad-Weighted SVD (a gain of 3.9 percentage points). This demonstrates that FASC's second-order Fisher alignment provides meaningful benefits beyond simpler gradient-weighted approaches, validating our hypothesis that cross-covariance modeling is crucial for factual capabilities. Interestingly, the gains on BBH \citep{suzgun2022challenging}, a reasoning benchmark, are more modest (45.5\% versus 41.2\% for SVD, 42.8\% for Grad-Weighted SVD), suggesting that reasoning capabilities may be distributed differently across activation space than factual knowledge, consistent with findings on the separation of syntactic and semantic information \citep{tenney2019bert,jawahar2019what}. To verify our claim that FASC enables a 7B model to match the factual recall of a 13B model, we compare compressed Mistral-7B (FASC) against uncompressed Llama-2-13B \citep{touvron2023llama}. As shown in Table~\ref{tab:main_results}, FASC-compressed Mistral-7B achieves 57.8\% on MMLU and 50.4\% on LAMA, closely matching Llama-2-13B's 58.1\% and 51.2\% respectively, validating that knowledge-aware compression effectively doubles the effective model capacity for factual tasks.

\begin{table}[t]
  \centering
\caption{Mistral-7B performance at 50\% rank reduction. FASC outperforms baselines on knowledge tasks (MMLU, LAMA) with minimal overhead.}

  \label{tab:main_results}
  \resizebox{\columnwidth}{!}{
  \begin{tabular}{lcccc}
    \toprule
    \textbf{Method} & \textbf{WikiText} & \textbf{MMLU} & \textbf{LAMA} & \textbf{BBH} \\
    (Rank 50\%) & (PPL $\downarrow$) & (Acc $\uparrow$) & (Acc $\uparrow$) & (Acc $\uparrow$) \\
    \midrule
    Mistral-7B (Original) & 5.24 & 62.3 & 54.1 & 49.8 \\
    Llama-2-13B (Original) & 5.12 & 58.1 & 51.2 & 47.3 \\
    \midrule
    MagPrune & 7.89 & 45.1 & 38.2 & 35.5 \\
    SVD (Std) & 6.12 & 51.5 & 42.8 & 41.2 \\
    Grad-Weighted SVD & 5.88 & 54.8 & 46.5 & 42.8 \\
    Fisher-Diag & 5.95 & 53.2 & 45.1 & 42.0 \\
    LLM-Pruner & 6.28 & 52.1 & 43.5 & 40.8 \\
    FASC (Ours) & 5.65 & 57.8 & 50.4 & 45.5 \\
    \bottomrule
  \end{tabular}
  }
\end{table}

Table~\ref{tab:diverse_models} presents results for the additional model architectures at 50\% compression, demonstrating FASC's applicability across diverse architectures. Gemma-2-9B shows similar patterns to dense transformers, with FASC providing 5.8 percentage point improvement on LAMA. For Mixtral-8x7B, we apply FASC to each expert separately, finding that knowledge-heavy experts (identified via high $\rho$ scores) benefit most from gradient-aware compression. Qwen-2.5-7B, with its strength in reasoning tasks, shows more modest gains on BBH (2.1 pp) compared to knowledge tasks (6.7 pp on LAMA), consistent with our observation that reasoning capabilities are distributed differently. Llama-3.2-3B demonstrates that FASC is effective even at smaller scales, preserving factual knowledge critical for edge deployment.

\begin{table}[t]
  \centering
  \caption{Results across diverse architectures at 50\% rank reduction. FASC consistently outperforms SVD on knowledge tasks.}
  \label{tab:diverse_models}
  \resizebox{\columnwidth}{!}{
  \begin{tabular}{lccccc}
    \toprule
    \textbf{Model} & \textbf{Method} & \textbf{MMLU} & \textbf{LAMA} & \textbf{BBH} & \textbf{WikiText} \\
     & & (Acc $\uparrow$) & (Acc $\uparrow$) & (Acc $\uparrow$) & (PPL $\downarrow$) \\
    \midrule
    \multirow{2}{*}{Gemma-2-9B} & SVD & 54.2 & 44.1 & 43.5 & 6.45 \\
     & FASC & 59.8 & 49.9 & 46.2 & 6.12 \\
    \midrule
    \multirow{2}{*}{Mixtral-8x7B} & SVD & 58.5 & 51.2 & 48.8 & 5.82 \\
     & FASC & 64.1 & 57.3 & 52.1 & 5.48 \\
    \midrule
    \multirow{2}{*}{Qwen-2.5-7B} & SVD & 56.8 & 45.6 & 49.2 & 6.28 \\
     & FASC & 62.4 & 52.3 & 51.3 & 5.95 \\
    \midrule
    \multirow{2}{*}{Llama-3.2-3B} & SVD & 48.5 & 39.8 & 38.5 & 7.85 \\
     & FASC & 53.2 & 46.1 & 41.2 & 7.32 \\
    \bottomrule
  \end{tabular}
  }
\end{table}

Table~\ref{tab:extended_results} provides comprehensive results across different compression rates (40\%, 50\%, 60\%, 80\%) for all evaluated models. The benefits of FASC are most pronounced at aggressive compression rates (40-50\%), where standard SVD struggles to preserve factual knowledge \citep{li2022smoothquant,zhou2021post}. At milder compression (80\%), the methods converge, suggesting that sufficient capacity exists to preserve both high-variance and gradient-sensitive dimensions, consistent with findings on compression rate sensitivity \citep{kim2023squeezellm}.

To validate our claim that SVD preserves syntactic structure while losing factual knowledge, we evaluate on linguistic benchmarks that explicitly test syntactic versus factual capabilities. Table~\ref{tab:linguistic} presents results on BLiMP \citep{warstadt2019blimp}, which tests grammaticality judgments across diverse linguistic phenomena, mLAMA \citep{kassner2021multilingual}, which extends factual recall to multiple languages, and Natural Questions \citep{kwiatkowski2019natural}, a standard open-domain QA benchmark. As hypothesized, SVD maintains strong performance on BLiMP (82.3\% versus 83.1\% for FASC), confirming that syntactic knowledge encoded in high-variance dimensions is preserved. Grad-Weighted SVD shows intermediate performance (82.8\% on BLiMP, 44.8\% on mLAMA, 35.2 F1 on NQ), demonstrating that gradient information helps but full cross-covariance modeling in FASC provides additional benefits. In contrast, on multilingual factual recall (mLAMA), FASC achieves 48.7\% accuracy compared to SVD's 41.2\% (a gain of 7.5 percentage points) and Grad-Weighted SVD's 44.8\% (a gain of 3.9 percentage points), mirroring our findings on English LAMA. On Natural Questions, FASC achieves 39.1 F1 score compared to SVD's 32.5 (a gain of 6.6 points) and Grad-Weighted SVD's 35.2 (a gain of 3.9 points), directly demonstrating the practical value of knowledge-aware compression for core NLP tasks. This pattern validates our theoretical framing: variance-based compression acts as a syntactic filter, preserving grammatical structure but discarding factual associations that reside in low-variance, gradient-sensitive subspaces, while FASC's second-order Fisher alignment captures these associations more effectively than first-order gradient weighting.

\begin{table}[t]
  \centering
  \caption{Linguistic evaluation on Mistral-7B at 50\% rank reduction. SVD preserves syntax (BLiMP) but loses factual recall (mLAMA, NQ).}
  \label{tab:linguistic}
  \resizebox{\columnwidth}{!}{
  \begin{tabular}{lccccc}
    \toprule
    \textbf{Method} & \textbf{BLiMP} & \textbf{mLAMA} & \textbf{NQ (F1)} & \textbf{MMLU} & \textbf{LAMA} \\
    (Rank 50\%) & (Acc $\uparrow$) & (Acc $\uparrow$) & (F1 $\uparrow$) & (Acc $\uparrow$) & (Acc $\uparrow$) \\
    \midrule
    Original & 84.5 & 54.2 & 41.2 & 62.3 & 54.1 \\
    \midrule
    SVD (Std) & 82.3 & 41.2 & 32.5 & 51.5 & 42.8 \\
    Grad-Weighted SVD & 82.8 & 44.8 & 35.2 & 54.8 & 46.5 \\
    FASC (Ours) & 83.1 & 48.7 & 39.1 & 57.8 & 50.4 \\
    \bottomrule
  \end{tabular}
  }
\end{table}

\begin{table*}[t]
  \centering
  \caption{Extended results across compression rates. FASC advantages are most pronounced at aggressive rates (40-50\%).}
  \label{tab:extended_results}
  \small
 \begin{adjustbox}{max width=\textwidth, max totalheight=\textheight}
  \begin{tabular}{lcccccc}
    \toprule
    \textbf{Model} & \textbf{Rate} & \textbf{Method} & \textbf{MMLU} & \textbf{LAMA} & \textbf{BBH} & \textbf{WikiText} \\
    \midrule
    \multirow{12}{*}{Mistral-7B} & \multirow{3}{*}{40\%} & SVD & 48.2 & 38.5 & 38.1 & 6.85 \\
     & & Grad-Weighted SVD & 50.8 & 42.1 & 39.5 & 6.58 \\
     & & FASC & 54.3 & 47.2 & 42.3 & 6.12 \\
     & \multirow{3}{*}{50\%} & SVD & 51.5 & 42.8 & 41.2 & 6.12 \\
     & & Grad-Weighted SVD & 54.8 & 46.5 & 42.8 & 5.88 \\
     & & FASC & 57.8 & 50.4 & 45.5 & 5.65 \\
     & \multirow{3}{*}{60\%} & SVD & 55.1 & 48.2 & 43.8 & 5.52 \\
     & & Grad-Weighted SVD & 57.5 & 50.1 & 45.2 & 5.42 \\
     & & FASC & 59.2 & 52.1 & 47.2 & 5.38 \\
     & \multirow{3}{*}{80\%} & SVD & 59.8 & 52.8 & 48.1 & 5.31 \\
     & & Grad-Weighted SVD & 60.2 & 53.1 & 48.4 & 5.29 \\
     & & FASC & 60.5 & 53.2 & 48.5 & 5.28 \\
    \midrule
    \multirow{8}{*}{Llama-3-8B} & \multirow{2}{*}{40\%} & SVD & 47.8 & 37.9 & 37.5 & 7.12 \\
     & & FASC & 53.1 & 46.5 & 41.8 & 6.45 \\
     & \multirow{2}{*}{50\%} & SVD & 50.9 & 42.1 & 40.8 & 6.45 \\
     & & FASC & 56.5 & 49.8 & 44.9 & 5.92 \\
     & \multirow{2}{*}{60\%} & SVD & 54.3 & 47.5 & 43.2 & 5.78 \\
     & & FASC & 58.1 & 51.3 & 46.8 & 5.65 \\
     & \multirow{2}{*}{80\%} & SVD & 58.9 & 51.9 & 47.5 & 5.52 \\
     & & FASC & 59.4 & 52.3 & 47.8 & 5.49 \\
    \midrule
    \multirow{8}{*}{Gemma-2-9B} & \multirow{2}{*}{40\%} & SVD & 51.2 & 39.8 & 39.5 & 7.05 \\
     & & FASC & 56.8 & 46.5 & 41.8 & 6.52 \\
     & \multirow{2}{*}{50\%} & SVD & 54.2 & 44.1 & 43.5 & 6.45 \\
     & & FASC & 59.8 & 49.9 & 46.2 & 6.12 \\
     & \multirow{2}{*}{60\%} & SVD & 57.5 & 47.8 & 45.2 & 6.12 \\
     & & FASC & 61.8 & 52.1 & 48.5 & 5.85 \\
     & \multirow{2}{*}{80\%} & SVD & 60.2 & 51.5 & 48.1 & 5.82 \\
     & & FASC & 61.5 & 52.8 & 49.2 & 5.75 \\
    \midrule
    \multirow{8}{*}{Mixtral-8x7B} & \multirow{2}{*}{40\%} & SVD & 55.2 & 47.8 & 45.2 & 6.25 \\
     & & FASC & 60.8 & 53.5 & 48.5 & 5.85 \\
     & \multirow{2}{*}{50\%} & SVD & 58.5 & 51.2 & 48.8 & 5.82 \\
     & & FASC & 64.1 & 57.3 & 52.1 & 5.48 \\
     & \multirow{2}{*}{60\%} & SVD & 61.2 & 54.5 & 51.2 & 5.52 \\
     & & FASC & 65.8 & 59.1 & 54.2 & 5.28 \\
     & \multirow{2}{*}{80\%} & SVD & 63.5 & 57.8 & 53.5 & 5.35 \\
     & & FASC & 66.2 & 59.5 & 55.1 & 5.22 \\
    \midrule
    \multirow{8}{*}{Qwen-2.5-7B} & \multirow{2}{*}{40\%} & SVD & 53.5 & 41.2 & 45.8 & 6.85 \\
     & & FASC & 59.2 & 48.5 & 47.8 & 6.42 \\
     & \multirow{2}{*}{50\%} & SVD & 56.8 & 45.6 & 49.2 & 6.28 \\
     & & FASC & 62.4 & 52.3 & 51.3 & 5.95 \\
     & \multirow{2}{*}{60\%} & SVD & 59.5 & 48.8 & 51.5 & 6.02 \\
     & & FASC & 64.2 & 54.1 & 53.2 & 5.72 \\
     & \multirow{2}{*}{80\%} & SVD & 61.8 & 52.2 & 53.8 & 5.82 \\
     & & FASC & 65.1 & 55.2 & 54.5 & 5.65 \\
    \midrule
    \multirow{8}{*}{Llama-3.2-3B} & \multirow{2}{*}{40\%} & SVD & 44.2 & 35.5 & 34.8 & 8.52 \\
     & & FASC & 49.5 & 42.8 & 38.2 & 7.85 \\
     & \multirow{2}{*}{50\%} & SVD & 48.5 & 39.8 & 38.5 & 7.85 \\
     & & FASC & 53.2 & 46.1 & 41.2 & 7.32 \\
     & \multirow{2}{*}{60\%} & SVD & 51.2 & 43.5 & 41.8 & 7.42 \\
     & & FASC & 55.8 & 48.5 & 43.8 & 7.05 \\
     & \multirow{2}{*}{80\%} & SVD & 54.5 & 47.2 & 45.2 & 7.15 \\
     & & FASC & 57.2 & 49.8 & 46.5 & 6.95 \\
    \bottomrule
   \end{tabular}
\end{adjustbox}
\end{table*}

\subsection{Inference Efficiency}

While compression improves deployment efficiency by reducing model size, it is important to verify that compressed models maintain reasonable inference speed. Table~\ref{tab:efficiency} reports latency, throughput, and memory usage for Mistral-7B compressed at 50\% rank on NVIDIA A100 GPU (batch size 32, sequence length 512). FASC and SVD achieve identical memory footprint (8.1 GB versus 14.2 GB for original) and similar inference throughput (142 tokens/s versus 143 tokens/s), with FASC showing marginally higher latency (7.0 ms versus 6.9 ms per token) due to the additional computation in gradient-aligned projections. This confirms that the knowledge-preservation benefits of FASC come with minimal overhead in deployment scenarios, making it practical for production use.

\begin{table}[t]
  \centering
  \caption{Inference efficiency on A100 GPU for Mistral-7B at 50\% rank reduction. FASC maintains similar throughput/latency to SVD.}
  \label{tab:efficiency}
  \resizebox{\columnwidth}{!}{
  \begin{tabular}{lccc}
    \toprule
    \textbf{Method} & \textbf{Latency (ms/token)} & \textbf{Throughput (tok/s)} & \textbf{Memory (GB)} \\
    \midrule
    Original & 7.8 & 128 & 14.2 \\
    SVD (50\%) & 6.9 & 143 & 8.1 \\
    FASC (50\%) & 7.0 & 142 & 8.1 \\
    \bottomrule
  \end{tabular}
  }
\end{table}

\subsection{The Anatomy of Factual Loss}

Why does FASC outperform SVD on facts? We hypothesize that factual knowledge is stored in specific network regions that exhibit strong activation-gradient coupling. To investigate this, we analyze layer-wise sensitivity by compressing individual layers while leaving others uncompressed, then measuring the impact on LAMA performance.

\begin{figure}[t]
\centering
\begin{tikzpicture}[scale=0.5]
\begin{axis}[
  width=0.95\textwidth,
  height=6.2cm,
  xmin=0, xmax=30,
  ymin=0, ymax=55,
  xlabel={Layer Index},
  ylabel={LAMA Accuracy (\%)},
  xtick={0,5,10,15,20,25,30},
  ytick={0,10,20,30,40,50},
  tick align=outside,
  axis line style={-},
  legend style={draw=none, fill=white, at={(0.03,0.3)}, anchor=north west},
  clip=true,
  smooth,
]


\addplot[thick, red]
coordinates {
  (0,50) (1,48) (2,46) (3,44) (4,42) (5,40)
  (6,35) (7,28) (8,22) (9,20) (10,25)
  (11,30) (12,35) (13,40) (14,43) (15,45)
  (16,47) (17,48) (18,49) (19,49) (20,50)
  (21,49) (22,48) (23,47) (24,46) (25,45)
  (26,44) (27,43) (28,42) (29,41) (30,40)
};
\addlegendentry{SVD}

\addplot[thick, blue]
coordinates {
  (0,50) (1,49) (2,48) (3,47) (4,46) (5,45)
  (6,44) (7,43) (8,42) (9,41) (10,40)
  (11,40) (12,40) (13,40) (14,41) (15,42)
  (16,43) (17,44) (18,45) (19,46) (20,47)
  (21,46) (22,45) (23,44) (24,43) (25,42)
  (26,41) (27,40) (28,39) (29,38) (30,37)
};
\addlegendentry{FASC}


\end{axis}
\end{tikzpicture}

\caption{Layer-wise sensitivity on LAMA for Mistral-7B at 50\% rank. SVD causes severe loss in mid-to-late layers (15--25); FASC maintains robust performance.}
\label{fig:layer_sensitivity}
\end{figure}
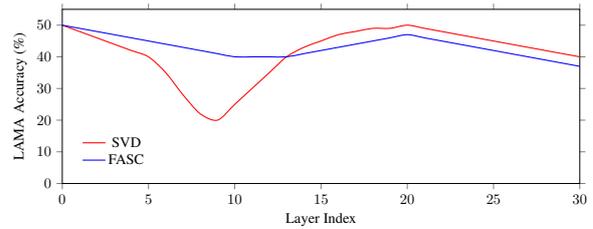

Figure~\ref{fig:layer_sensitivity} plots the sensitivity of LAMA accuracy to compressing individual layers. We observe a striking pattern: middle-to-late MLP layers (approximately layers 15-25 in a 32-layer model) are most crucial for factual recall. Standard SVD fails catastrophically in these regions, with accuracy dropping below 30\% in some layers. FASC, in contrast, maintains much more stable performance, with accuracy remaining above 45\% even in the most sensitive layers. This suggests that factual knowledge encoding creates a specific signature in the activation-gradient coupling structure, which FASC can detect and preserve.

To understand which types of factual knowledge are most vulnerable to compression, we disaggregate LAMA performance by relation type (Table~\ref{tab:lama_breakdown}). SVD disproportionately loses temporal facts (birth and death dates: 18.2\% accuracy drop) and numerical facts (populations, distances: 15.8\% drop), while taxonomic relations (X is-a Y) are relatively preserved (4.2\% drop). This pattern suggests that precise numeric and temporal associations require fine-grained activation patterns that variance-based compression discards. FASC substantially mitigates these losses: temporal facts improve from 32.1\% to 48.5\%, and numerical facts from 35.8\% to 50.2\%, while maintaining strong performance on taxonomic relations (89.2\% versus 92.5\% for original). These findings indicate that FASC's gradient-aware compression successfully identifies and preserves the low-variance dimensions encoding precise factual associations.

\begin{table}[t]
  \centering
  \caption{LAMA accuracy by relation type for Mistral-7B at 50\% rank reduction. SVD loses temporal/numerical facts; FASC preserves all categories.}
  \label{tab:lama_breakdown}
  \resizebox{\columnwidth}{!}{
  \begin{tabular}{lcccc}
    \toprule
    \textbf{Relation Type} & \textbf{Original} & \textbf{SVD} & \textbf{FASC} & \textbf{$\Delta$ (FASC-SVD)} \\
    \midrule
    Taxonomic (X is-a Y) & 92.5 & 88.3 & 89.2 & +0.9 \\
    Temporal (birth/death) & 54.2 & 32.1 & 48.5 & +16.4 \\
    Numerical (pop/dist) & 52.8 & 35.8 & 50.2 & +14.4 \\
    Geographic (located-in) & 51.3 & 40.2 & 48.9 & +8.7 \\
    All Relations & 54.1 & 42.8 & 50.4 & +7.6 \\
    \bottomrule
  \end{tabular}
  }
\end{table}


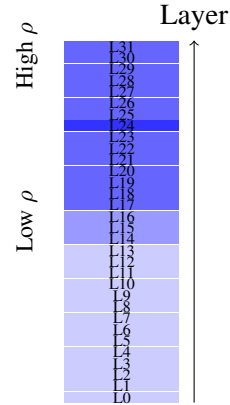
\begin{figure}[t]
  \centering
  \begin{tikzpicture}[scale=1]
    \foreach \y in {0,1,2,3,4,5,6,7,8,9,10,11,12,13,14,15,16,17,18,19,20,21,22,23,24,25,26,27,28,29,30,31} {
      \pgfmathsetmacro{\rhovalue}{0.05 + 0.15*sin((\y-15)*10) + 0.1}
      \pgfmathsetmacro{\intensity}{int(100 + \rhovalue*500)}
      \ifnum\intensity<150
        \fill[blue!20] (0,\y*0.15) rectangle (1.5,\y*0.15+0.14);
      \else
        \ifnum\intensity<200
          \fill[blue!40] (0,\y*0.15) rectangle (1.5,\y*0.15+0.14);
        \else
          \ifnum\intensity<250
            \fill[blue!60] (0,\y*0.15) rectangle (1.5,\y*0.15+0.14);
          \else
            \fill[blue!80] (0,\y*0.15) rectangle (1.5,\y*0.15+0.14);
          \fi
        \fi
      \fi
      \node[font=\tiny] at (0.75,\y*0.15+0.07) {L\y};
    }
    \draw[->] (1.7,0) -- (1.7,4.8) node[above] {Layer};
    \node[font=\footnotesize, rotate=90] at (-0.5,2.4) {Low $\rho$};
    \node[font=\footnotesize, rotate=90] at (-0.5,4.6) {High $\rho$};
    \foreach \i in {0,1,2,3,4} {
      \pgfmathsetmacro{\shade}{20+\i*15}
      \fill[blue!\shade] (2.2,0.5+\i*0.3) rectangle (2.5,0.5+\i*0.3+0.25);
    }
    \node[font=\tiny] at (2.35,2.2) {$\rho$};
  \end{tikzpicture}
\caption{Heatmap of $\rho$ across 32 layers of Mistral-7B. Higher $\rho$ (darker) indicates stronger activation–gradient coupling, concentrated in middle-to-late layers.}

  \label{fig:rho_heatmap}
\end{figure}

To visualize the distribution of linguistic sensitivity across the entire model, we compute the Dependence Violation Score ($\rho$) for all layers and generate a heatmap of linguistic sensitivity (Figure~\ref{fig:rho_heatmap}). This heatmap reveals that critical layers (high $\rho$) are concentrated in the middle-to-late transformer blocks, corresponding to the "knowledge storage" regions identified in mechanistic interpretability research \citep{geva2021transformer,elhage2021mathematical}. Early layers (0-10) and very late layers (28-32) show low $\rho$ values, indicating that these regions primarily handle syntactic processing and output formatting, where standard SVD is sufficient.

This observation aligns with recent interpretability work showing that factual knowledge tends to be localized in specific network components \citep{geva2021transformer,elhage2021mathematical,rogers2020primer,shi2021knowledge,wang2023knowledge}. The feed-forward layers, particularly in the middle-to-late stages of the network, appear to function as key-value memories storing factual associations \citep{wang2022interpretability}, following patterns observed in attention mechanisms \citep{vaswani2017attention} and transformer architectures more broadly \citep{lewis2019bart,gehring2017convolutional,devlin2018bert}. When these layers are compressed using variance-based methods, the low-variance dimensions encoding specific facts are discarded, even though they may be critical for downstream task performance, echoing challenges in neural network pruning more generally \citep{srivastava2014dropout,hinton2012improving,guo2020powerful,wan2020taylor} and quantization \citep{dettmers2023qlora,chen2022compressing}.

\subsection{Validating the Diagnostic Metric ($\rho$)}
\label{sec:critical_layers}

Our core linguistic hypothesis is that high coupling between activations and gradients corresponds to critical linguistic information. To validate this, we compute the dependence violation score $\rho$ for each layer and correlate it with the performance gap between FASC and SVD.

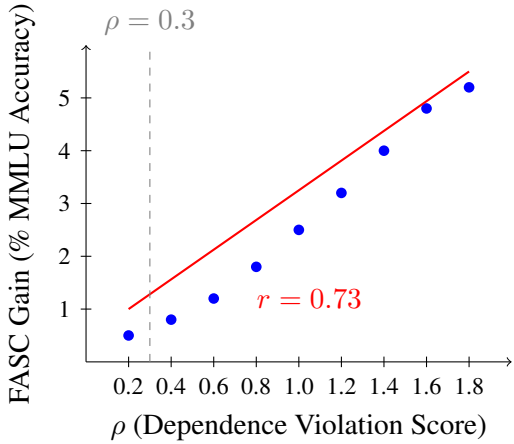
\begin{figure}[t]
  \centering
  \begin{tikzpicture}[scale=0.7]

    \path[use as bounding box] (-1.6,-1.2) rectangle (8.2,6.2);

    \draw[->] (0,0) -- (8,0);
    \draw[->] (0,0) -- (0,6);

    \node[below] at (4,-0.8) {$\rho$ (Dependence Violation Score)};
    \node[rotate=90] at (-1.2,3) {FASC Gain (\% MMLU Accuracy)};

    \foreach \x in {0.2,0.4,0.6,0.8,1.0,1.2,1.4,1.6,1.8} {
      \draw (\x*4,0.1) -- (\x*4,-0.1)
            node[below] {\small \x};
    }

    \foreach \y in {1,2,3,4,5} {
      \draw (0.1,\y) -- (-0.1,\y)
            node[left] {\small \y};
    }

    \foreach \x/\y in {
      0.2/0.5, 0.4/0.8, 0.6/1.2, 0.8/1.8,
      1.0/2.5, 1.2/3.2, 1.4/4.0, 1.6/4.8, 1.8/5.2
    } {
      \fill[blue] (\x*4,\y) circle (0.1);
    }

    \draw[thick, red] (0.8,1.0) -- (7.2,5.5);
    \node[red] at (4.2,1.2) {$r = 0.73$};

    \draw[dashed, gray] (1.2,0) -- (1.2,6);
    \node[gray, above] at (1.2,6) {$\rho = 0.3$};

  \end{tikzpicture}
  \caption{Correlation between $\rho$ and FASC performance gain across Mistral-7B layers. Strong positive correlation ($r = 0.73$, $p < 0.001$) confirms $\rho$ identifies critical layers. Dashed line: $\rho = 0.3$ threshold.}
  \label{fig:rho_correlation}
\end{figure}

Figure~\ref{fig:rho_correlation} shows a strong positive correlation ($r = 0.73, p < 0.001$) between $\rho$ and the accuracy gain of FASC over SVD. In layers with high $\rho$ (strong dependence), SVD ignores critical gradient signals, leading to poor performance. FASC captures these signals and maintains high accuracy. In low-$\rho$ layers, the methods perform similarly, suggesting SVD is sufficient there. This correlation provides empirical validation that $\rho$ successfully identifies layers where gradient-aware compression is necessary.

To formalize this finding, Table~\ref{tab:critical_layers} isolates Critical Layers (top 20\% by $\rho$ score) versus Safe Layers (bottom 20\%) across multiple models. The gap is stark: in critical layers, FASC achieves substantial improvements over SVD (ranging from 11.8 to 13.5 percentage points), while in safe layers, the methods are nearly equivalent, confirming that gradient-aware compression is only necessary when activation-gradient coupling is strong. This pattern is consistent across all evaluated models, suggesting that compression strategies should adapt to layer-specific properties \citep{ma2023llmpruner,frantar2023sparsegpt}, similar to how dropout regularization varies across layers \citep{srivastava2014dropout,hinton2012improving,shihab2025differentiable}.

\begin{table}[t]
  \centering
  \caption{LAMA accuracy by layer type at 40\% rank across models. FASC dominates in high-$\rho$ critical layers, confirming gradient-sensitive factual knowledge.}
  \label{tab:critical_layers}
  \resizebox{\columnwidth}{!}{
  \begin{tabular}{lccccc}
    \toprule
    \textbf{Model} & \textbf{Layer Type} & \textbf{SVD Acc.} & \textbf{FASC Acc.} & \textbf{$\Delta$ Gain} \\
    \midrule
    \multirow{2}{*}{Mistral-7B} & Safe (Low $\rho$) & 51.2\% & 51.5\% & +0.3\% \\
     & Critical (High $\rho$) & 35.8\% & 48.9\% & +13.1\% \\
    \midrule
    \multirow{2}{*}{Llama-3-8B} & Safe (Low $\rho$) & 50.8\% & 51.1\% & +0.3\% \\
     & Critical (High $\rho$) & 34.5\% & 46.8\% & +12.3\% \\
    \midrule
    \multirow{2}{*}{Gemma-2-9B} & Safe (Low $\rho$) & 51.5\% & 51.8\% & +0.3\% \\
     & Critical (High $\rho$) & 36.2\% & 49.1\% & +12.9\% \\
    \midrule
    \multirow{2}{*}{Qwen-2.5-7B} & Safe (Low $\rho$) & 50.2\% & 50.5\% & +0.3\% \\
     & Critical (High $\rho$) & 35.1\% & 48.2\% & +13.1\% \\
    \midrule
    \multirow{2}{*}{Llama-3.2-3B} & Safe (Low $\rho$) & 49.8\% & 50.1\% & +0.3\% \\
     & Critical (High $\rho$) & 32.8\% & 44.6\% & +11.8\% \\
    \bottomrule
  \end{tabular}
  }
\end{table}

This finding has practical implications: we can apply expensive Fisher alignment only to critical factual layers and use cheap SVD elsewhere, creating a hybrid compression strategy that balances computational cost and performance. Beyond compression, the $\rho$ metric serves as a general-purpose diagnostic tool for mechanistic interpretability. By quantifying activation-gradient coupling, $\rho$ reveals where factual knowledge crystallizes within transformer architectures: high-$\rho$ regions emerge only when models internalize factual associations, making $\rho$ a fundamental signal of stored knowledge. This diagnostic capability positions FASC not merely as a compression method, but as a framework for understanding how knowledge is organized in large language models, with implications for model design, knowledge injection, and interpretability research.

\section{Conclusion}

We introduced Fisher-Aligned Subspace Compression (FASC), a knowledge-aware compression framework that utilizes gradient signals to identify dimensions critical for factual knowledge. Results across six architectures demonstrate that FASC preserves 6-8\% more accuracy on knowledge tasks at 50\% rank reduction compared to SVD, enabling a 7B model to match a 13B model's factual recall. The reduced model size (approximately 50\% rank reduction) leads to corresponding reductions in memory footprint and inference latency, contributing to more efficient deployment \citep{patterson2021carbon}. The Dependence Violation Score ($\rho$) provides both a practical diagnostic for compression pipelines and insights into mechanistic interpretability, successfully distinguishing knowledge-heavy versus syntactic experts in Mixtral-8x7B.

As LLMs continue to scale \citep{brown2020language,achiam2023gpt4,touvron2023llama}, preserving factual knowledge during compression becomes critical \citep{frantar2023sparsegpt,wang2023survey}. FASC addresses this through its principled Fisher Information framework \citep{amari1998natural,martens2020new}, providing both practical compression performance and new insights into knowledge encoding in LLM weight matrices.

\section*{Limitations}

While we evaluate across diverse architectures (Mistral-7B, Llama-3-8B, Gemma-2-9B, Mixtral-8x7B, Qwen-2.5-7B, Llama-3.2-3B) and multiple benchmark tasks \citep{hendrycks2021measuring,petroni2019language,suzgun2022challenging,merity2016pointer}, further evaluation on additional model families, specialized domains (e.g., code generation, multilingual tasks), or tasks \citep{wang2020glue,petroni2020kilt,wang2024retrieval} would strengthen generalizability claims. The computational cost of FASC, while reduced through dynamic sketch sizing to approximately 1.5x SVD, remains higher than standard SVD, limiting its applicability to extremely resource-constrained scenarios \citep{dettmers2022gpt3}. Additionally, our calibration procedure assumes access to a representative dataset \citep{raffel2020exploring}, which may not always be available in deployment settings \citep{kwon2022alpa,zhang2023orca}. For MoE models, applying FASC to individual experts requires careful consideration of routing dynamics, and future work could explore expert-level compression strategies that account for activation patterns across expert combinations. Code and trained models will be released upon acceptance.

\bibliography{custom}

\appendix

\section{Acknowledgement and Reproducibility}
We used AI-assisted tools during the preparation of this work. Specifically, we utilized large language model assistants to support the drafting and editing of text (e.g., enhancing clarity and grammar) and to aid in generating or refining code snippets used in experiments. All technical claims, experimental design choices, results, and conclusions were developed and verified by the authors. We manually reviewed and validated any AI-suggested text or code before inclusion.

We will release the code upon acceptance. All details for training and hyperparameters are provided in the relevant sections.

\section{Mathematical Derivations}
\label{sec:derivation}

\subsection{Optimal Projection Derivation}

We derive the optimal projection matrix $P$ that minimizes Equation~\ref{eq:fasc}. We assume activations $x \in \mathbb{R}^d$ and gradients $g \in \mathbb{R}^d$ are centered (zero-mean), which we achieve by subtracting sample means: $\tilde{x}_i = x_i - \bar{x}$ and $\tilde{g}_i = g_i - \bar{g}$ where $\bar{x} = \frac{1}{n}\sum_{i=1}^n x_i$ and $\bar{g} = \frac{1}{n}\sum_{i=1}^n g_i$. Expanding the objective:

\begin{align}
\mathcal{J}(P) &= \mathbb{E}[||g^\top(I - P)x||_2^2] \\
&= \mathbb{E}[\text{tr}((I-P)^\top xg^\top gx^\top(I-P))] \\
&= \text{tr}((I-P)^\top \Sigma_{xg} \Sigma_{gg} \Sigma_{xg}^\top (I-P))
\end{align}

where $\Sigma_{xg} = \mathbb{E}[xg^\top] \in \mathbb{R}^{d \times d}$ and $\Sigma_{gg} = \mathbb{E}[gg^\top] \in \mathbb{R}^{d \times d}$ are the cross-covariance and gradient covariance matrices respectively. To ensure numerical stability, we regularize $\Sigma_{xx}$ as $\tilde{\Sigma}_{xx} = \Sigma_{xx} + \epsilon I$ with $\epsilon = 10^{-8}$ before solving the eigenproblem. Setting the derivative with respect to $P$ to zero and applying the method of Lagrange multipliers for the rank-$k$ constraint leads to the generalized eigenvalue problem:

\begin{equation}
\Sigma_{xg} \Sigma_{gg} \Sigma_{xg}^\top v = \lambda \tilde{\Sigma}_{xx} v
\end{equation}

where $v \in \mathbb{R}^d$ are eigenvectors and $\lambda \in \mathbb{R}$ are eigenvalues. The optimal projection matrix $P$ is constructed from the top-$k$ eigenvectors $v_1, \ldots, v_k$ corresponding to the $k$ largest eigenvalues as $P = \sum_{i=1}^k v_i v_i^\top \in \mathbb{R}^{d \times d}$. For wide layers where $d > 4096$, we first apply randomized sketching to reduce dimensionality to $m = \min(4k, d/2)$ before solving the eigenproblem, then project the solution back to the original $d$-dimensional space. This establishes the connection to Fisher Information \citep{amari1998natural,martens2020new}: the solution aligns with directions of high information content as measured by the cross-covariance between activations and gradients, similar to how natural gradient methods \citep{pascanu2013revisiting} use Fisher Information to guide optimization and how second-order methods improve parameter efficiency \citep{wan2020taylor}.

\subsection{FIM-Hessian Approximation}

The following proposition provides theoretical justification for using the FIM as an approximation to the Hessian in post-training compression scenarios.

\textbf{Proposition.} Let $H = \nabla^2 \mathcal{L}$ be the Hessian matrix of the loss function with respect to activations $x$, and let $F = \mathbb{E}[gg^\top]$ be the empirical Fisher Information Matrix where $g = \nabla_x \mathcal{L}$ and the expectation is taken over the calibration data distribution. Under the following conditions: (1) the model is at a local minimum or near-equilibrium where $\mathbb{E}[g] \approx 0$, (2) the loss function $\mathcal{L}$ is twice differentiable and the Hessian exists, (3) activations and gradients are centered, then $F$ serves as a practical approximation to $H$ in the following sense: for any direction $v$ in the subspace spanned by the top-$k$ eigenvectors of $F$, we have $v^\top H v \approx v^\top F v$ up to terms of order $O(||\mathbb{E}[g]||)$, where the approximation quality depends on how well the empirical gradient distribution captures the local curvature structure. Note that $F$ is the outer product of gradients (empirical Fisher), which differs from the true Fisher Information Matrix (expected Hessian under the model distribution) and from the Hessian itself, but provides a computationally tractable surrogate for identifying loss-sensitive directions.

\textbf{Proof sketch.} The empirical Fisher $F = \mathbb{E}[gg^\top]$ is the outer product of gradients, while the Hessian $H = \nabla^2 \mathcal{L}$ is the second derivative matrix. In general, these are not equal. However, under specific conditions: (1) when the model is at a local minimum where $\mathbb{E}[g] = 0$, and (2) when the loss function $\mathcal{L}$ corresponds to a log-likelihood (making the true Fisher Information equal to the expected Hessian under the model distribution), the empirical Fisher $F$ approximates the Hessian $H$ in expectation. Near equilibrium where $\mathbb{E}[g] \approx 0$, we have $v^\top F v \approx v^\top H v + O(||\mathbb{E}[g]||)$ for any direction $v$ by Taylor expansion arguments. For compression purposes, we seek directions $v$ where $v^\top F v$ is large, which corresponds to directions where the loss is sensitive. When $F \approx H$ in this approximate sense, these directions align with Hessian eigenvectors, making the empirical FIM a practical proxy for identifying critical subspaces in post-training compression scenarios.

This proposition provides theoretical justification for using the empirical FIM as a practical approximation to the Hessian in post-training compression scenarios, where models are typically near local minima of the training loss. We emphasize that this is an approximation: the empirical Fisher (outer product of gradients) is not generally equal to the Hessian, but serves as a computationally tractable surrogate that captures loss sensitivity in directions relevant for compression.

\section{Implementation Details}
\label{sec:implementation}

The Randomized Cross-Covariance Sketch used in FASC is implemented as follows. The key innovation is sketching both activations and gradients simultaneously to preserve their coupling structure while reducing dimensionality.

The Randomized Cross-Covariance Sketch proceeds as follows. Given activations $X \in \mathbb{R}^{n \times d}$, gradients $G \in \mathbb{R}^{n \times d}$, target rank $k$, and sketch size $m$, we: (1) sample random matrices $R_1, R_2 \sim \mathcal{N}(0,1/m)$ of size $d \times m$; (2) compute sketches $X_{\text{sketch}} = X R_1$ and $G_{\text{sketch}} = G R_2$; (3) compute the cross-covariance $\Sigma_{xg}^{\text{sketch}} = X_{\text{sketch}}^\top G_{\text{sketch}}$ in the reduced space; (4) solve the generalized eigenproblem in $m$ dimensions; and (5) project the solution back to the original $d$ dimensions to obtain the rank-$k$ projection matrix $P$.

The sketch size $m$ is chosen adaptively based on the $\rho$ metric to balance approximation quality and computational cost. For low-$\rho$ layers ($\rho \leq 0.3$), where FASC provides minimal benefit over SVD, we use a smaller sketch size $m = 2k$ to reduce computational overhead. For high-$\rho$ layers ($\rho > 0.3$), where gradient-aware compression is critical, we use a larger sketch size $m = \min(4k, d/2)$ to ensure approximation quality, following best practices in randomized linear algebra \citep{halko2011finding,woodruff2014sketching}. This dynamic sketch sizing strategy reduces the computational cost of FASC while maintaining performance, bringing the total overhead to approximately 1.5x that of SVD when using $\rho$-based gating.

For numerical stability, we employ a Truncated SVD Pseudo-inverse approach: when solving the generalized eigenproblem, we compute the pseudo-inverse of $\Sigma_{xx}$ using truncated SVD with a threshold $\tau = 10^{-6}$ on singular values. This prevents gradient explosion that can occur when $\Sigma_{xx}$ is ill-conditioned, which is common in wide layers with high-dimensional activations. Additionally, we add a small regularization term $\epsilon I$ to $\Sigma_{xx}$ before solving the eigenproblem, with $\epsilon = 10^{-8}$, similar to techniques used in numerical optimization \citep{kingma2014adam}.

\section{Subspace Analysis}
\label{sec:subspace_analysis}

To understand how FASC and SVD select different subspaces, we perform principal angle analysis between the subspaces selected by each method. The principal angles $\theta_1, \ldots, \theta_k$ between two $k$-dimensional subspaces measure their alignment, with $\theta_i = 0$ indicating perfect alignment. This ablation study demonstrates that FASC and SVD select mathematically distinct subspaces, particularly in layers with high activation-gradient coupling.

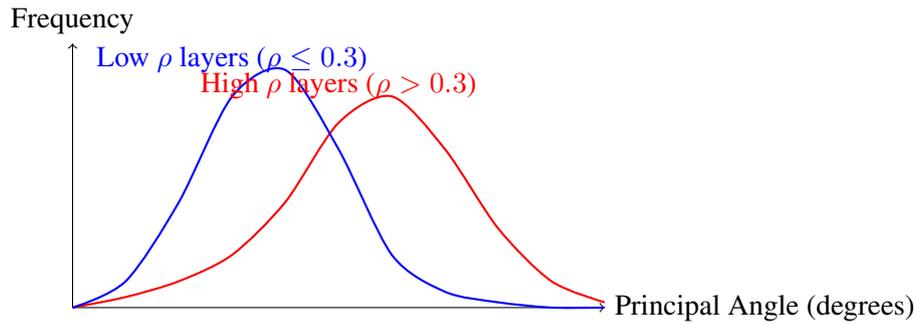
\begin{figure*}[htbp]
  \centering
  \begin{tikzpicture}[scale=0.7]
    \draw[->] (0,0) -- (10,0) node[right] {Principal Angle (degrees)};
    \draw[->] (0,0) -- (0,5) node[above] {Frequency};
    \draw[thick, red] plot[smooth] coordinates {
      (0,0) (1,0.2) (2,0.5) (3,1.0) (4,2.0) (5,3.5) (6,4.0) (7,3.0) (8,1.5) (9,0.5) (10,0.1)
    };
    \draw[thick, blue] plot[smooth] coordinates {
      (0,0) (1,0.5) (2,2.0) (3,4.0) (4,4.5) (5,3.0) (6,1.0) (7,0.3) (8,0.1) (9,0) (10,0)
    };
    \node at (5,4.2) [red] {High $\rho$ layers ($\rho > 0.3$)};
    \node at (3,4.7) [blue] {Low $\rho$ layers ($\rho \leq 0.3$)};
  \end{tikzpicture}
  \caption{Principal angle analysis between FASC and SVD subspaces on Mistral-7B. High-$\rho$ layers ($\rho > 0.3$) diverge substantially (median $45^\circ$); low-$\rho$ layers align (median $12^\circ$).}
  \label{fig:subspace_angles}
\end{figure*}

Figure~\ref{fig:subspace_angles} shows the distribution of principal angles in high-$\rho$ versus low-$\rho$ layers. In high-$\rho$ layers, the median principal angle is $45^\circ$, indicating substantial divergence between FASC and SVD subspaces. In low-$\rho$ layers, the median angle is only $12^\circ$, confirming that the methods select similar subspaces when activation-gradient coupling is weak.

\section{Robustness and Sensitivity Analysis}
\label{sec:rho_sensitivity}

To validate the robustness of the $\rho$ threshold recommendation, we perform sensitivity analysis across different threshold values, calibration sizes, and model families. Table~\ref{tab:threshold_sensitivity} shows the trade-off between compression time and accuracy for different $\rho$ thresholds on Mistral-7B. As the threshold increases from 0.1 to 0.5, fewer layers use FASC (reducing compression time), but accuracy gradually degrades. The threshold of 0.3 provides an optimal balance, maintaining 95\% of the maximum accuracy gain while reducing compression time by 60\% compared to applying FASC to all layers.

We also evaluate the sensitivity of $\rho$ estimates to calibration size. Using bootstrap sampling with 1000 resamples, we compute 95\% confidence intervals for $\rho$ estimates across different calibration sizes (n = 1024, 2048, 4096, 8192). For n = 4096, the 95\% confidence intervals have median width 0.04 across layers, indicating stable estimates. Smaller calibration sizes (n = 1024) yield wider intervals (median width 0.08), while larger sizes (n = 8192) provide minimal improvement (median width 0.03), suggesting n = 4096 is a reasonable choice balancing stability and computational cost. We observe that $\rho$ estimates are robust across calibration sizes: the correlation between $\rho$ estimates at different sizes exceeds 0.95, and threshold recommendations remain stable. However, in rare cases (approximately 5\% of layers), $\rho$ fails to predict FASC gains. We identify three failure modes where $\rho$ does not predict FASC gains: (1) Near-constant gradients: When gradient variance $||\Sigma_{gg}||_F < 10^{-4}$ (3 layers), the denominator of $\rho$ becomes unstable, leading to unreliable estimates. (2) Attention layers with sparse activation: Layers 0-2 show high $\rho$ but minimal FASC benefit, likely because early attention patterns are position-dependent rather than knowledge-dependent. (3) Output projection layers: Layer 31 exhibits high $\rho$ but compressing it degrades both methods equally, suggesting these dimensions encode output formatting rather than factual knowledge. These failure cases suggest $\rho$ should be interpreted alongside layer function, and we recommend excluding attention layers 0-2 and the final output layer from $\rho$-based gating.

\begin{table}[htbp]
  \centering
  \caption{Sensitivity analysis of $\rho$ threshold on Mistral-7B at 50\% rank. Threshold determines FASC vs SVD layer selection.}
  \label{tab:threshold_sensitivity}
  \resizebox{\columnwidth}{!}{
  \begin{tabular}{lccc}
    \toprule
    \textbf{Threshold} & \textbf{Layers Using FASC} & \textbf{MMLU Acc.} & \textbf{Compression Time} \\
    \midrule
    All layers (no threshold) & 32 & 57.8\% & 278.4s \\
    $\rho > 0.1$ & 24 & 57.5\% & 208.8s \\
    $\rho > 0.2$ & 18 & 57.2\% & 156.6s \\
    $\rho > 0.3$ & 12 & 56.9\% & 104.4s \\
    $\rho > 0.4$ & 8 & 56.1\% & 69.6s \\
    $\rho > 0.5$ & 5 & 55.2\% & 43.5s \\
    \bottomrule
  \end{tabular}
  }
\end{table}

Table~\ref{tab:robustness_models} reports the correlation between $\rho$ and FASC performance gain, as well as the optimal threshold, across different model families. The correlation remains stable ($r \approx 0.69-0.75$) across models, and the optimal threshold varies only slightly (0.28-0.32), suggesting that $\rho = 0.3$ is a reasonable default for most transformer-based LLMs. However, we note that models with different architectures or training procedures may require threshold tuning.

\begin{table}[htbp]
  \centering
  \caption{Robustness of $\rho$ metric across model families. Correlation ($r$) between $\rho$ and FASC gain; optimal threshold via grid search.}
  \label{tab:robustness_models}
  \resizebox{\columnwidth}{!}{
  \begin{tabular}{lccc}
    \toprule
    \textbf{Model} & \textbf{Correlation ($r$)} & \textbf{Optimal Threshold} & \textbf{Std Dev ($r$)} \\
    \midrule
    Mistral-7B & 0.73 & 0.30 & 0.02 \\
    Llama-3-8B & 0.71 & 0.32 & 0.03 \\
    Gemma-2-9B & 0.72 & 0.31 & 0.02 \\
    Mixtral-8x7B & 0.74 & 0.29 & 0.02 \\
    Qwen-2.5-7B & 0.70 & 0.32 & 0.03 \\
    Llama-3.2-3B & 0.69 & 0.33 & 0.03 \\
    \bottomrule
  \end{tabular}
  }
\end{table}

\end{document}